\documentclass[runningheads]{llncs}

\usepackage[mobile]{eccv}

\usepackage{eccvabbrv}

\usepackage[pdftex]{graphicx}
\usepackage{booktabs} 
\usepackage{lipsum}
\usepackage{graphicx}
\usepackage{comment}
\usepackage{colortbl}
\usepackage{wrapfig}
\usepackage{mathtools}
\usepackage{wrapfig}
\usepackage{float}
\usepackage{caption}
\usepackage{multirow}
\usepackage[final]{pdfpages}

\usepackage[accsupp]{axessibility}  

\usepackage[pagebackref,breaklinks,colorlinks,citecolor=eccvblue]{hyperref}

\usepackage{orcidlink}

\definecolor{lightyellow}{rgb}{1,1, 0.8}
\definecolor{tabgreen}{HTML}{E6FFE3}
\definecolor{tabred}{HTML}{FFEEEE}
\definecolor{tabyellow}{HTML}{FFFEE5}
\definecolor{tabgold}{HTML}{d4ae36}
\definecolor{tabsilver}{HTML}{c0c0c0}
\definecolor{tabbronze}{HTML}{967344}
\definecolor{red}{rgb}{1, 0.7, 0.7} 
\definecolor{orange}{rgb}{1, 0.85, 0.7} 
\definecolor{yellow}{rgb}{1, 1, 0.7} 

\begin{document}

\title{Flying with Photons:\\Rendering Novel Views of Propagating Light}
\titlerunning{Flying with Photons}

\author{Anagh Malik$^{1, 2}$\thanks{anagh@cs.toronto.edu} \quad
Noah Juravsky$^{1}$ \quad
Ryan Po$^{3}$ \\
Gordon Wetzstein$^{3}$ \quad
Kiriakos N. Kutulakos$^{1, 2}$ \quad
David B. Lindell$^{1, 2}$ \vspace{-0.5em}}

\authorrunning{A. Malik et al.}

\institute{$^1$University of Toronto \quad $^2$Vector Institute \quad $^3$Stanford University \\ {\small\href{https://anaghmalik.com/FlyingWithPhotons}{anaghmalik.com/FlyingWithPhotons}} \vspace{-1.5em}} 

\maketitle

\begin{abstract}

We present an imaging and neural rendering technique that seeks to synthesize videos of light propagating through a scene from novel, moving camera viewpoints.
Our approach relies on a new ultrafast imaging setup to capture a first-of-its kind, multi-viewpoint video dataset with picosecond-level temporal resolution.
Combined with this dataset, we introduce an efficient neural volume rendering framework based on the \textit{transient field}. 
This field is defined as a mapping from a 3D point and 2D direction to a high-dimensional, discrete-time signal that represents time-varying radiance at ultrafast timescales. 
Rendering with transient fields naturally accounts for effects due to the finite speed of light, including viewpoint-dependent appearance changes caused by light propagation delays to the camera. We render a range of complex  effects, including scattering, specular reflection, refraction, and diffraction. Additionally, we demonstrate removing viewpoint-dependent propagation delays using a time warping procedure, rendering of relativistic effects, and video synthesis of direct and global components of light transport.

\end{abstract}

\vspace{-2em}
\section{Introduction}
\label{sec:intro}

\begin{figure}[h!]
    \centering
    \includegraphics[width=\textwidth]{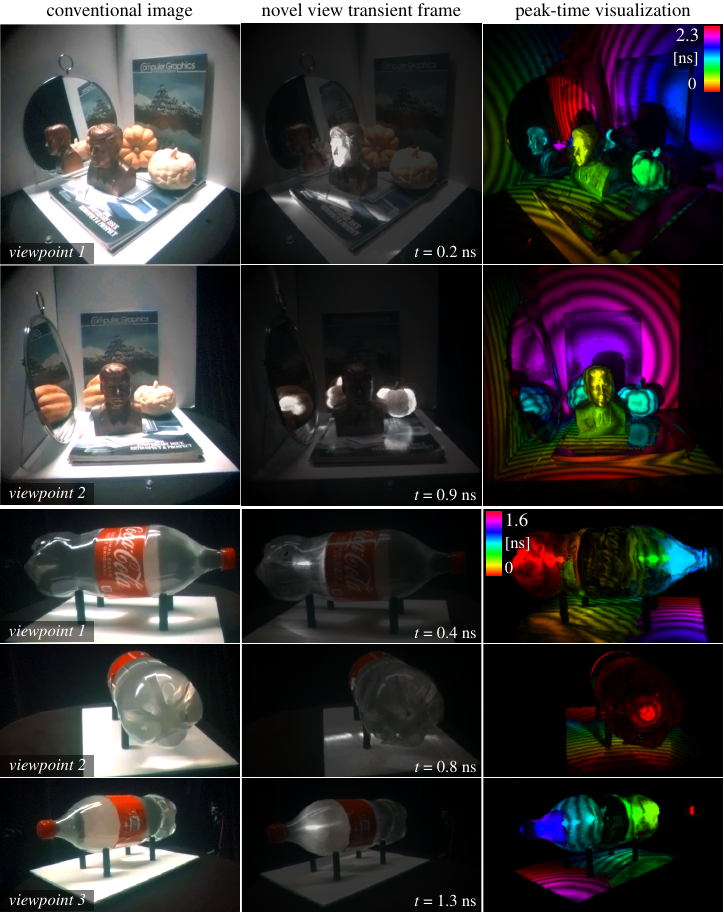}
    \caption{\textbf{Flying with Photons.} The input to our method is a set of multi-viewpoint transient videos that capture a scene illuminated by a diffused or collimated pulsed light source. We then render videos of propagating light---\textit{transient videos}---from different novel viewpoints at different moments in time.
    \textbf{Left:} Conventional image of the \textit{Kennedy} and \textit{Coke bottle} scenes. 
\textbf{Centre:} Grayscale transient frames rendered from novel viewpoints, composited over the colour image of the scene.  
\textbf{Right:} Adapting the peak-time visualization of Velten et al.~\cite{velten2013femto}, we show the entire transient video in a single frame. 
Hue indicates the time at which the peak intensity is observed at each pixel, and brightness corresponds to the magnitude of the peak intensity. We modulate the brightness over the time dimension to create color bands corresponding to equi-time paths (a.k.a \textit{isochrones}), which reveal the shape of propagating wavefronts.}

\label{fig:teaser}
\end{figure}

By imaging at trillions of frames per second, ultrafast cameras record videos of propagating light.
These \textit{transient videos} reveal the appearance of the world at the speed of light~\cite{velten2013femto}.
Such measurements of light transport are useful in a variety of applications: they can be used to understand fundamental mechanisms in physics~\cite{liang2018single}, to recover material properties~\cite{naik2011single}, and to image through living tissue~\cite{pifferi2016new} or behind occluders~\cite{faccio2020non}. 
Our specific aim is to synthesize transient videos from arbitrary, dynamic camera viewpoints (Fig.~\ref{fig:teaser}), enabling flexible visualization of light transport and facilitating new applications based on 3D representations of light propagation.

Current state-of-the-art techniques for novel view synthesis use images of a scene captured from multiple viewpoints~\cite{tewari2022advances}.
However, these techniques are at odds with existing setups for transient videography that are primarily intended for single-viewpoint capture.
For example, existing transient video systems use interferometry~\cite{abramson1978light,gkioulekas2015micron,kotwal2023passive} or pulsed lasers combined with streak cameras~\cite{gao2014single,velten2013femto} or single-photon avalanche diodes (SPADs)~\cite{o2017reconstructing,gariepy2015single}.
Due to the complexity of these capture setups, multi-viewpoint transient videos have so far been available only in simulation~\cite{jarabo2017recent}, and exploring applications in novel view synthesis has been precluded by a lack of captured datasets.

\clearpage

Applying existing novel view and video synthesis techniques~\cite{fridovich2023k,cao2023hexplane,wang2023mixed,li2022neural,mildenhall2021nerf} to multi-view transient rendering is not straightforward because they are designed for timescales that are too long for the speed of light to matter. 
In our setting, the timescale is short enough that the distance between the camera and the scene---and the corresponding speed-of-light time delay---dramatically affects the measurements. 
Thus, rendering at novel camera--scene distances requires special treatment, and the camera's velocity can matter as well (e.g., inducing relativistic effects~\cite{jarabo2015relativistic}).

Our approach models the finite speed of light and synthesizes transient videos from dynamic (or static) novel viewpoints. 
We demonstrate the approach using a first-of-its-kind dataset of multi-viewpoint transient videos, captured with a gantry that enables precise azimuth and elevation control of a scanning, single-pixel SPAD (i.e., on the hemisphere surrounding a scene).
The SPAD records photon arrivals with picosecond-level accuracy, and is synchronized to a diffused or collimated illumination source that is stationary with respect to the scene and emits picosecond-scale pulses at megahertz rates.

To render transient videos from novel viewpoints, we introduce a volumetric representation that is optimized based on multiple input transient videos, each captured with a static camera from a different viewpoint.  
Although we consider rendering at speed-of-light timescales, the static cameras of our dataset do not capture relativistic effects---we consider simulating these effects separately as an extension of our method (Sec.~\ref{sec:applications}). 

Our transient rendering procedure adapts a neural representation~\cite{muller2022instant,li2023nerfacc} to learn two fields---a conventional density field~\cite{mildenhall2021nerf} and \textit{transient field}: a mapping from a 3D point and 2D direction to a high-dimensional, discrete-time signal that represents time-varying radiance at ultrafast timescales.
After learning these fields, we sample the transient field and density field at points along a camera ray and apply volume rendering to these samples to obtain the synthesized transient waveform at a camera pixel.

Our rendering procedure also accounts for light propagation delays to each camera's centre of projection. Specifically, we apply a ``time unwarping'' procedure similar to Velten et al.~\cite{velten2013femto}; applied in our context, it allows us to learn a canonical spacetime representation of the scene, which can then be time-warped to account for propagation delays to any camera viewpoint.

In summary, we provide the following contributions.
\begin{itemize}
    \item We render light propagating through real scenes from novel, moving viewpoints for the first time.
    \item We introduce a transient field representation to make this possible and evaluate our method on a range of scenes with inter-reflections, multiple scattering, refraction, and diffraction.
    \item We build a system for multi-viewpoint transient videography and capture a dataset of scenes with complex light transport effects.
    \item We show additional capabilities, including time unwarping, relativistic rendering, and direct--global separation of light transport.
\end{itemize}

\section{Related Work}
\label{sec:related_work}

\paragraph{Capturing light transport.} Our work is closely related to time-resolved imaging modalities that directly capture light in flight~\cite{velten2013femto,o2017reconstructing,lindell2018towards}.
Among these modalities, interferometry provides the highest temporal resolution and resolves light propagation at micron scales~\cite{gkioulekas2015micron,kotwal2023passive}.
Still, interferometric techniques typically require bulky optical setups and have a limited working range.
Streak cameras achieve sub-picosecond resolution~\cite{wang2020single}, but are also bulky and expensive, making them more difficult to use.
On the other hand, photonic mixer devices are far less expensive and can be used to reconstruct transient videos, though at coarser, nanosecond scales~\cite{heide2013low,o2014temporal}.
SPADs provide a middle ground between these sensing modalities, as they are relatively inexpensive, yet have high temporal resolution at the picosecond level~\cite{piron2020review}. 
Our work uses a scanned, single-pixel SPAD with $\approx$50 ps temporal resolution.

Various other methods have been developed to capture phenomena relating to light transport.
For example, structured illumination can separate light into different components (e.g., direct or indirect) by projecting high frequency patterns~\cite{nayar2006fast,gupta2011stuctured,mirdehghan2024turbosl}, illuminating and imaging based on epipolar lines~\cite{o20143d,o2015homogeneous}, or using a combination of structured illumination and masking~\cite{o2012primal}. 
Other work captures light paths directly by imaging scenes placed into a fluorescent medium~\cite{hullin2008direct}.
However, none of these techniques can capture propagating light.

\paragraph{Transient rendering.} Transient renderers model light transport effects that arise from the finite speed of light~\cite{smith2008rendering,jarabo2014framework}.
They account for propagation delays between light sources and surfaces in the scene~\cite{pediredla2019ellipsoidal,royo2022non}, as well as effects due to refraction~\cite{ament2014radiative}, scattering~\cite{jarabo2018bidirectional}, birefringence~\cite{jarabo2014framework}, or volume absorption and scattering~\cite{pediredla2020path}. 
Differentiable transient renderers optimize scene parameters using transient measurements~\cite{yi2021differentiable,wu2021differentiable}. 
While they have shown promise for non-line-of-sight imaging~\cite{iseringhausen2020non,tsai2019beyond}, scaling these techniques to multi-view captures of complex scenes remains a challenge. 

At transient timescales, camera motion induces relativistic effects, including time dilation, distortions that arise from the contraction of spacetime (Lorentz contraction), Doppler shifts, and the searchlight effect~\cite{jarabo2015relativistic,weiskopf1999relativity,weiskopf2006explanatory}. 
Our dataset does not capture relativistic effects because the camera is static during capture. 
Thus, our primary aim is not to achieve physically accurate rendering of such effects; however, we do include a limited extension to our approach that simulates certain relativistic effects, inspired by the method of Jarabo et al.~\cite{jarabo2015relativistic}.

Similar to us, Jarabo et al.~\cite{jarabo2015relativistic} also explore rendering transient videos from novel viewpoints. 
However, their approach uses a single-viewpoint transient video for which the scene geometry is known. 
They extract a textured mesh by projecting the transient video appearance onto the scene geometry, and this enables re-rendering from any viewpoint.  
Our approach is significantly more general as we jointly optimize for geometry and the view-dependent appearance of transient videos, and we incorporate data from multiple viewpoints.

We are also inspired by Velten et al.~\cite{velten2013femto}; we use a similar style of peak-time visualization (shown in Fig.~\ref{fig:teaser}), and we extend their time warping procedure to model propagation delays in our volume rendering framework.
In their approach, time warping was used to remove the scene--camera path length in a single-viewpoint transient video.
In our work, time warping appears as a propagation delay that we add to samples of the transient field along a camera ray. 
We also demonstrate using time warping to remove scene--camera path lengths in transient videos rendered with dynamic cameras, which removes view-dependent propagation delays and makes the visualization more intuitive.

\paragraph{Neural rendering.} Our work builds on methods for 3D reconstruction and novel view synthesis based on neural radiance fields (NeRFs)~\cite{mildenhall2021nerf}.
While such techniques have been developed to, e.g., model refractive objects~\cite{bemana2022refractive,pan2022refractive,zhan2023nerfrac} and glossy surfaces~\cite{verbin2022refnerf}, or to perform inverse rendering tasks~\cite{zhang2021nerfactor,mai2023neural,jin2023tenoir,klinghoffer2023plato,mu2024towards,shen2021non}, no previous method performs novel view synthesis of complex, time-resolved light transport effects observed from multiple viewpoints.
To do so requires a dataset that captures these effects and a representation to facilitate rendering them. Neither of which are currently available; our method provides both.

Closest to our work is TransientNeRF~\cite{malik2023transient}, which performs novel view synthesis of the direct component of transient measurements, i.e., from the data used in lidar. 
However, TransientNeRF cannot render indirect light transport effects, and relies on a lidar-specific image formation model, where the sensor and light source are coaxial.
In contrast, our approach handles non-coaxial light sources, such as a point source or collimated beam placed anywhere within a scene.
Further, we render more general light transport effects, such as diffuse and specular reflections, multiple scattering, refraction, and diffraction.

\section{Rendering Propagating Light with Transient Fields}
\label{sec:method}

\subsection{SPAD Measurement Model} 

Consider a scene that is illuminated by an ultrashort pulse of light from a stationary laser source that is diffused or collimated. 
The wavefront of this impulse interacts with the scene through a potentially complex sequence of events, including reflection, refraction, or scattering before being reflected back to a camera pixel along a ray $\mathbf{r}$.
We model the time-varying photon radiance of light along $\mathbf{r}$ as the impulse response of the scene $h(\mathbf{r}, t)$.
Then, the expected number of photons $\lambda_\mathbf{r}$ collected by an ideal detector during a time bin of width $W$ is proportional to the integral of photon radiance over time: $\lambda_\mathbf{r}[n] \propto \int_{nW}^{(n+1)W} h(\mathbf{r}, t)\ \mathrm{d}t$. 

\begin{figure*}[t!]
    \includegraphics{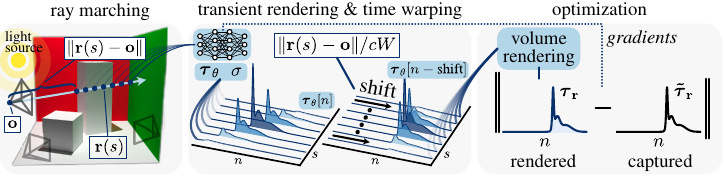}
\caption{We cast a ray into the scene and query a neural representation at samples $\mathbf{r}(s)$ along the ray to retrieve an $N$-dimensional transient $\boldsymbol{\tau}_\theta$ and density value $\sigma$ for each sample.
Each element $\boldsymbol{\tau}_\theta[n]$ of that transient corresponds to a time bin of width $W$ times the speed of light $c$. We time-shift each transient based on the time delay from the camera origin $\mathbf{o}$ to $\mathbf{r}(s)$ and composite them together using volume rendering. The neural representation parameters are optimized to minimize the difference between the rendered transient $\boldsymbol{\tau}_\mathbf{r}$ and the captured transient $\boldsymbol{\tilde{\tau}}_\mathbf{r}$.}
    \label{fig:method}
    \vspace{-2em}
\end{figure*}

In practice, we use a single-photon avalanche diode (SPAD) to count the number of photons detected within each time bin, resulting in the measured transient $\boldsymbol{\tilde{\tau}}_\mathbf{r}$. 
The photon detections  are distributed according to an inhomogeneous Poisson process whose time-varying rate function is $\lambda_\mathbf{r}$~\cite{rapp2017few}:
\begin{equation}
    \boldsymbol{\tilde{\tau}}_\mathbf{r}[n] \sim \textsc{Poisson}(P \eta \lambda_\mathbf{r}[n] + B), \quad B = P(\eta A_\mathbf{r} + D).
\end{equation}
Here, $P$ represents the number of laser pulse periods used to capture measurements, $\eta$ is the detection efficiency of the SPAD, and $B$ is the expected number of photons due to ambient light $A_\mathbf{r}$ and the dark count rate $D$ of the sensor.
Although SPADs exhibit other second-order effects, such as dead time, afterpulsing, or cross-talk~\cite{bronzi2015spad}, these can be neglected in the low-flux regime considered in this work, where less than 5\% of emitted laser pulses lead to a photon detection~\cite{oconnor1984time}.
Increasing the number of laser pulses results in improved signal-to-noise ratio and measurements that better approximate the photon arrival rate.

\subsection{Transient Fields and Rendering} 
We use a volumetric representation of the scene to render the transient $\boldsymbol{\tau}_\mathbf{r}\in\mathbb{R}_+^N$ given a camera ray $\mathbf{r}$ (see Fig.~\ref{fig:method}). 
Specifically, we define a point on a ray $\mathbf{r}(s) = \mathbf{o} + s\mathbf{d}$ where $\mathbf{o}\in\mathbb{R}^3$ is the camera center of projection and $\mathbf{d}$ is a three-dimensional unit vector corresponding to the ray direction. 
The transient field, $\boldsymbol{\tau}_\theta : (\mathbf{r}(s), \mathbf{d}) \mapsto \mathbb{R}_+^N$, is a neural representation with parameters $\theta$ that maps a three-dimensional point and a ray direction to the discrete time bin values of a transient. 
Additionally, the neural representation outputs a spatially varying absorption coefficient $\sigma(\mathbf{r}(s))$, which gives the differential probability of ray termination at each point in the volume.
Finally, the transient along ray $\mathbf{r}$ is computed using a modified version of the volume rendering equation~\cite{attal2021torf,gkioulekas2016evaluation}: 
\begin{align}
    \boldsymbol{\tau}_\mathbf{r} &= \int_{s_n}^{s_f} T(s)\sigma(\mathbf{r}(s))\left[\boldsymbol{\tau}_\theta(\mathbf{r}(s), \mathbf{d})\ast \underbrace{\delta\left[n - \lVert \mathbf{r}(s) - \mathbf{o} \rVert / (cW)\right]}_{\text{propagation delay}}\right] \ \mathrm{d}s,\nonumber
\\
    \text{where } T(s) &= \exp\left(-\int_{s_n}^s \sigma(\mathbf{r}(u))\ \mathrm{d} u \right).
    \label{eq:render}
\end{align}
Here, $T$ represents transmittance from ray distance $s_n$ to $s$, and $c$ is the speed of light. 
The transient vectors $\boldsymbol{\tau}_\theta$ sampled along the ray are time shifted via convolution ($\ast$) with the Kronecker delta function $\delta$---this accounts for the propagation delay to the camera and is similar to the time warping procedure of Velten et al.~\cite{velten2013femto}.
If the time shift is not modeled, there is an ambiguous mapping from $(\mathbf{r}(s), \mathbf{d})$ to shifted versions of the same transient, depending on the distance to the camera center of projection.
In practice we apply a continuous shift using linear interpolation. 
The volume rendering integral is evaluated numerically using the quadrature rule proposed by Max~\cite{max1995optical}.

\subsection{Implementation Details}
\paragraph{Optimization procedure.}
We optimize the transient field representation $\boldsymbol{\tau}_\theta$ using SPAD measurements $\boldsymbol{\tilde{\tau}}^{(v)}_\mathbf{r}$ captured from $0 \leq v \leq V-1$ different viewpoints.
The neural representation is optimized using the loss function
\begin{equation}
    \mathcal{L} = \sum\limits_{v, \mathbf{r}, n} \lVert g(\boldsymbol{\tilde{\tau}}_{\mathbf{r}}^{(v)}[n]) - \boldsymbol{\tau}_{\mathbf{r}}^{(v)}[n]   \rVert_2^2,
    \label{eq:loss}
\end{equation}
where the summation is over all viewpoints, camera rays, and transient time bins.
Since the measured transients can have a high dynamic range (e.g., spanning zero to thousands of photons), we apply a gamma function, $g(x) = x^{1/\gamma}$, to compress the dynamic range and to improve the ability of the neural representation to fit weak signals, such as multiply scattered light.

We parameterize $\boldsymbol{\tau}_\theta$ using an adapted version of the NerfAcc~\cite{li2022nerfacc} implementation of Instant-NGP~\cite{muller2022instant} (see supplement).
The model is trained using the Adam optimizer~\cite{kingma2015adam} until convergence, which typically occurs after 500k iterations for the simulated dataset and 1M iterations for the captured dataset.
During training, we anneal the learning rate after 50\%, 75\%, and 90\% of the training iterations by a factor of 0.33.
Training requires roughly 10 hours for the simulated scenes and 20 hours for the captured scenes due to the higher temporal resolution of transient in the captured data. We select the batch size to fit within 48 GB of VRAM on an NVIDIA A40 GPU.
We use $\gamma=5$ for simulated measurements and $\gamma=2$ for captured measurements, which we set empirically to balance contrast and detail in the rendered transient videos.
To render color transients for the simulated dataset, we modify the transient field such that $\boldsymbol{\tau}_\theta : (\mathbf{r}(s), \mathbf{d}) \mapsto \mathbb{R}_+^{3N}$, i.e., it outputs a separate transient for each color channel.
We assume that the camera intrinsics and extrinsics are known, and we describe our calibration procedure in the following section.

\paragraph{Dynamic viewpoint rendering.} After training, the same rendering procedure can be used to create transient videos of a scene from a dynamic camera viewpoint. 
This is accomplished by defining a time-varying camera trajectory consisting of camera extrinsics, $[\mathbf{R}_n|\mathbf{t}_n]$ (i.e., rotation and translation), corresponding to every transient time bin $\boldsymbol{\tau}_\mathbf{r}[n]$.
The $n$th frame in the output transient video is rendered by computing $\boldsymbol{\tau}_\mathbf{r}[n]$ for the camera rays transformed by the extrinsics.

\section{Multi-Viewpoint Transient Dataset}

\paragraph{Simulated dataset.}
We use the renderer of Liu et al.~\cite{liu2022temporally} as well as a version of Mitsuba 2 modified for transient rendering~\cite{royo2022non} to simulate transient videos from multiple viewpoints.
The illumination is a point source that emits a temporal impulse into the scene. For the Mitsuba 2 scenes we move the camera to sample a grid of 31 by 3 viewpoints on a hemisphere spanning 45 to 180 degrees in azimuth and 30 to 45 degrees in elevation.
We create a total of 4 synthetic scenes based on the Cornell box or assets from Blendswap (\url{https://blendswap.com/}) and Bitterli~\cite{bitterli16resources}, which we assemble together in Blender~\cite{blender}.
To evaluate the method, we render a set of 30 unseen viewpoints sampled on the same hemisphere as the training views, which is consistent with what we capture using our hardware system.
We also render 30 additional unseen viewpoints by sampling on a hemisphere whose radius is roughly $20\%$ greater than the one used for training.
The simulated transients are rendered at 512$\times$512 resolution and the number of time bins ranges from 300 to 900 depending on the scene.

\paragraph{Hardware prototype.}

\begin{wrapfigure}[18]{r}{0.5\textwidth}
    \vspace{-2em}
    \includegraphics[width=0.5\textwidth]{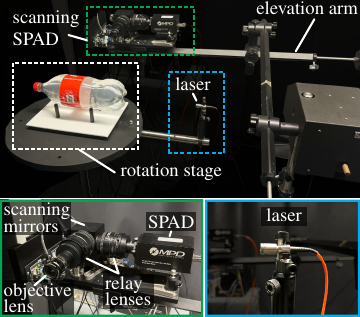}
    \caption{Multi-viewpoint capture setup. A gantry controls the elevation angle of a scanning SPAD, and a scene and laser source are rotated together on a stage.}
    \label{fig:hardware}
\end{wrapfigure}

We capture a multi-viewpoint transient video dataset using a prototype system (Fig.~\ref{fig:hardware}) comprising a single-pixel SPAD, a 2D scanning galvonometer, a set of relay lenses, and an objective lens. 
The SPAD is synchronized to a 532 nm laser that emits 35 ps pulses of light at a 10 MHz repetition rate. 
A time-correlated single-photon counter time stamps photon arrivals with a system resolution of approximately 70 ps, and the scanning galvonometer is configured to raster scan the scene at 512$\times$512 resolution. 
We couple the free-space laser beam into a multi-mode fiber and set the output power to keep the incident photon flux to roughly 500k counts per second on average, low enough to avoid pileup.  

The camera gantry consists of a rotation stage and an elevation arm on which we mount the SPAD and scanning system.
The emission end of the multi-mode fiber and captured scene are mounted on the rotation stage so that the illumination source and scene move together.
To capture a conventional color image of the scene with our system, we turn the room lights on and perform sequential captures with red, green, and blue color filters placed in front of the objective lens.

\paragraph{Calibration and captured dataset.}
The camera intrinsics are calibrated by capturing a sequence of intensity images of a checkerboard and using the MATLAB camera calibration toolbox~\cite{zhang2000flexible,matlabcamera}. 
To calibrate the extrinsics, we place a textured scene and checkerboard on the rotation stage and capture an image from each viewpoint used to capture transient videos.
We use COLMAP~\cite{schonberger2016structure} to solve for the extrinsics, and we scale the resulting camera translations so that the reconstructed size of the checkerboard squares matches the known geometry.
We use the same camera intrinsics and extrinsics for all captured scenes.

The captured dataset consists of 5 scenes, and for each scene we capture 45 (or 75) transient videos on a grid of 15 (or 25) by 3 viewpoints spanning 125 to 360 degrees in azimuth and 15 degrees in elevation.
Capturing each transient video requires 20 to 30 minutes; we bin all captured photons into a transient histogram with 4096 bins, each spanning 4 ps.
See the supplement for a detailed description of capture parameters and measured photon counts for each scene.
Prior to using the simulated and captured datasets for view synthesis with our method, we normalize them as described in the supplement to ensure that the dynamic range is compatible with the output range of the neural representation.
All code and datasets are publicly available on the project webpage.

\section{Results}
\label{sec:results}

We evaluate the proposed method for rendering propagating light from novel viewpoints in simulation and using captured datasets. 
We compare to baselines, including a modified version of Transient NeRF (T-NeRF)~\cite{malik2023transient}; while this method is intended for lidar view synthesis, we adapt its ray marching scheme to handle the non-coaxial point light source and camera used in our simulated scenes (detailed in the supplement). 
However, we omit T-NeRF from our evaluation on captured data because it requires the light source position (which we do not calibrate), and we show that it cannot model global effects in the simulated results.
We also compare to K-Planes~\cite{fridovich2023k}, a method for video novel view synthesis that uses a spatiotemporal feature grid and a volume rendering model. 
Finally, we compare two versions of the proposed method, one with and one without explicitly modeling the propagation delay to the camera (Eq.~\ref{eq:render}).

\begin{figure*}[h!]
    \centering
    \includegraphics[width=0.85\textwidth]{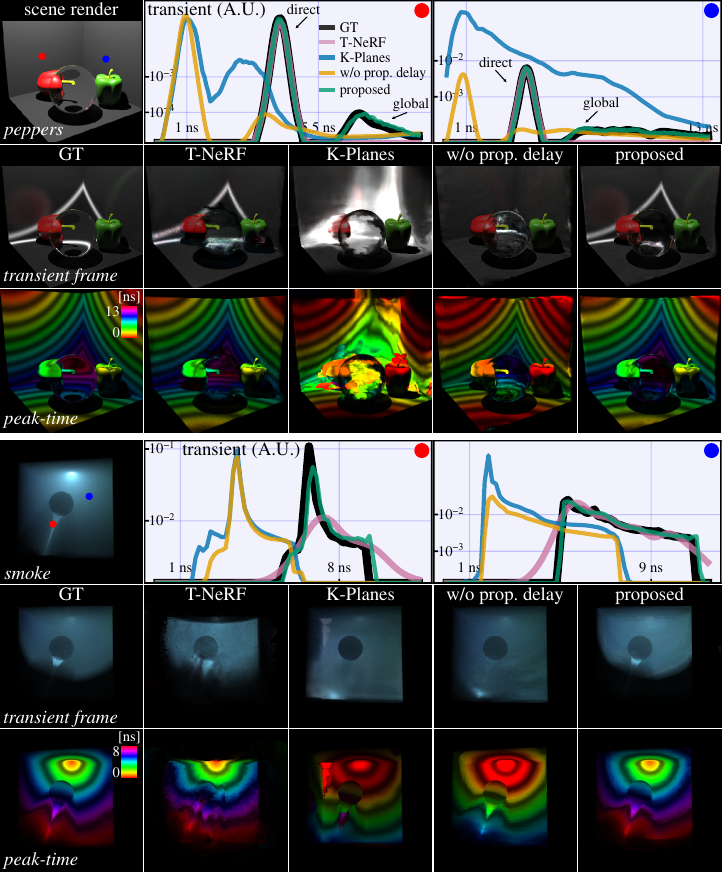}
    \captionof{figure}{Simulated results rendered from novel viewpoints for the \textit{peppers} and \textit{smoke} scenes. \
    \textbf{Rows 1, 4:} The ground-truth integrated transient is shown alongside transient plots for all methods. The proposed method more accurately represents both the direct and global components of light. \
    \textbf{Rows 2, 5:} For all methods, we show one frame of the transient video, composited over the integrated image of the scene.
    \textbf{Rows 3, 6:} Peak-time visualization illustrating the transient in a single frame. Hue encodes the time of peak intensity, brightness is modulated by the maximum intensity, and each band corresponds to an isochrone, or wavefront of equal path length.}
    \label{fig:sims}
    \vspace{-2em}
\end{figure*}

\begin{table}[t!]
    \captionof{table}{Evaluation of transient rendering from novel viewpoints. We report parameters and time to render a transient video (NVIDIA A6000).}
    \label{tab:results}
    \centering
    \begin{tabular}{llcccccc}
        \toprule
        & method & param. & time & PSNR (dB)$\,\uparrow$ & LPIPS$\,\downarrow$ & SSIM$\,\uparrow$ & T-IOU$\,\uparrow$ \\\midrule
        \parbox[t]{6mm}{\multirow{4}{*}{\rotatebox[origin=c]{90}{\textit{simulated}}}} & 
        T-NeRF~\cite{malik2023transient} & 15M & \cellcolor{red}7.1s & \cellcolor{yellow}26.349 &
        \cellcolor{yellow}0.338 &
        \cellcolor{orange}0.887&
        \cellcolor{orange}0.729  \\
        & K-Planes~\cite{fridovich2023k} & 37M & 320.7s & 20.551 &
        0.431&
        0.666&
        \cellcolor{yellow}0.358 \\
        & (w/o prop.\ delay) & 15M & \cellcolor{orange}11.9s & \cellcolor{orange}27.791 &
        \cellcolor{orange}0.334&
        \cellcolor{yellow}0.861&
        0.334  \\
        & proposed & 15M & \cellcolor{yellow}12.8s & \cellcolor{red}32.965 &
        \cellcolor{red}0.247&
        \cellcolor{red}0.965&
        \cellcolor{red}0.830 \\
        \toprule
        \parbox[t]{6mm}{\multirow{3}{*}{\rotatebox[origin=c]{90}{\textit{captured}}}}& 
        K-Planes~\cite{fridovich2023k} & 43M & \cellcolor{yellow}37min &
        \cellcolor{orange}24.115 &
        \cellcolor{orange}0.516&
        \cellcolor{orange}0.594&
        \cellcolor{orange}0.395  \\
        & (w/o prop.\ delay) & 15M & \cellcolor{red}5.78s & \cellcolor{yellow}17.118 &
        \cellcolor{yellow}0.529&
        \cellcolor{yellow}0.346&
        \cellcolor{yellow}0.174 \\
        & proposed & 15M & \cellcolor{orange}28.0s & \cellcolor{red}24.949 &
        \cellcolor{red}0.431&
        \cellcolor{red}0.666&
        \cellcolor{red}0.468 \\\bottomrule
    \end{tabular}
\vspace{-1em}
\end{table}

\subsection{Simulated Transient Rendering}

We evaluate the approach in simulation on four different scenes: \textit{pots}, \textit{peppers}, \textit{Cornell box}, and \textit{smoke}.
We render all scenes using a modified version of Mitsuba 2~\cite{royo2022non}, except for \textit{smoke}, for which we use the method of Liu et al.~\cite{liu2022temporally} because it supports rendering participating media. 
We show rendered novel views and transients on held-out camera viewpoints for the proposed method and baselines in Fig.~\ref{fig:sims}.
The proposed method outperforms the baselines in terms of accuracy and computational efficiency.
Since T-NeRF cannot model indirect effects, it only recovers the direct component of light transport.
K-Planes models time-varying illumination, but does not account for propagation delays to different camera viewpoints (Eq.~\ref{eq:render}), and so fails to learn the view-dependent shifts in the time-varying illumination. 
We observe a similar performance degradation in the proposed method without propagation delay modeling; thus, accurate propagation delay modeling is key to accurate view synthesis for this task.

We report image quality and accuracy of synthesized transients from held-out camera viewpoints, averaged across all scenes, in Table~\ref{tab:results}. 
Prior to evaluating our method, we undo the gamma correction learned during training (Eq.~\ref{eq:loss}).
To evaluate image quality, we average the rendered transient videos over the time dimension, normalize, gamma correct ($\gamma=2.2$), and compute PSNR, LPIPS~\cite{zhang2018unreasonable}, and SSIM~\cite{wang2004image}.
We assess the quality of synthesized transients by introducing a transient intersection over union (IoU) metric. 
This is calculated as 
\begin{equation}
\textsc{IoU}(\boldsymbol{\tau}_1, \boldsymbol{\tau}_2) = \frac{\sum_n \min\left(\boldsymbol{\tau}_1[n], \boldsymbol{\tau}_2[n]\right)}{\sum_n \max\left(\boldsymbol{\tau}_1[n], \boldsymbol{\tau}_2[n]\right)},
\end{equation}
and we report the average transient IoU across all scenes.
See the supplement for implementation details of the metrics and for the individual scene metrics. 

To evaluate computational efficiency, we measure the time it takes for each method to render a transient video from a single viewpoint; we also report the number of parameters in each model (see Table~\ref{tab:results}). 
For this test, we use an NVIDIA A6000 GPU and set the batch size for each method to use the maximum 48 GB of VRAM. 
K-Planes requires roughly 25$\times$ longer than both T-NeRF and the proposed method because each time bin requires its own rendering pass.

\begin{figure*}[t!]
    \centering
    \vspace{-1em}
    \includegraphics[width=0.85\textwidth]{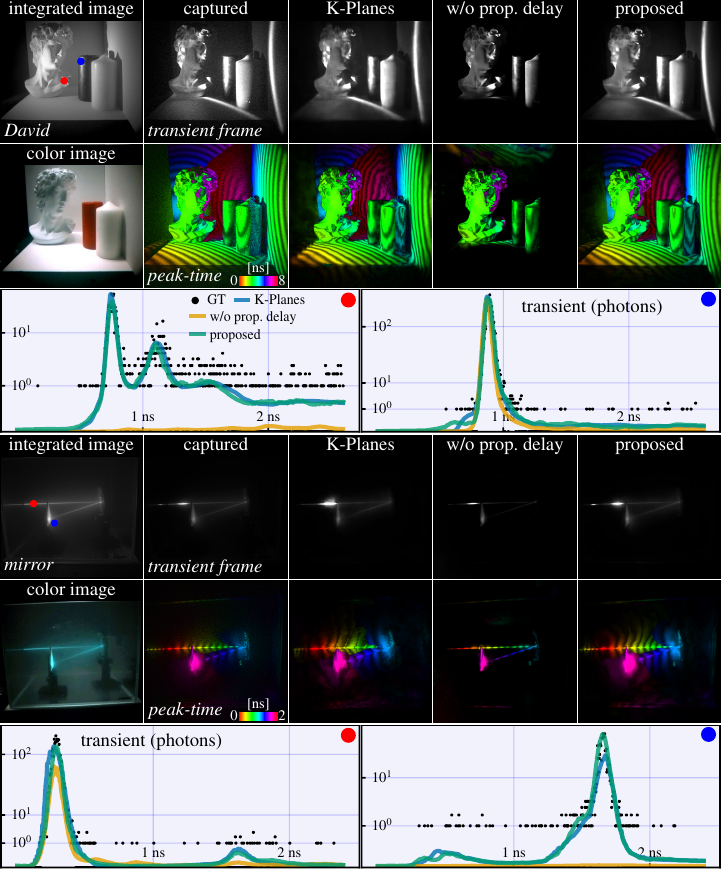}
    \captionof{figure}{Captured results rendered from novel viewpoints for the \textit{David} and \textit{mirror} scenes. \
    \textbf{Row 1, 4:} For all methods, we show one frame of the transient video, composited over the integrated image of the scene.
    \textbf{Row 2, 5:} Peak-time visualization. 
    \textbf{Row 3, 6:} Transient plots. We show the captured photon count using a scatter plot instead of a continuous line, due to the sparse and quantized nature of the measurements. 
    The methods reconstruct a continuous approximation of the underlying transient and suppress the noise observed in the captured data.}
    \label{fig:capture}
    \vspace{-2em}
\end{figure*}

\subsection{Transient Rendering with Captured Data}

We capture five different scenes to evaluate our method and compare with baseline approaches.
The \textit{Coke bottle} scene is shown in Fig.~\ref{fig:teaser} and is similar to the result shown by Velten et al.~\cite{velten2013femto}, except we perform multi-viewpoint reconstruction. 
We use a collimated beam to illuminate a Coca-Cola bottle filled with water and a small amount of milk.
To illuminate the \textit{Kennedy} (Fig.~\ref{fig:teaser}) and  \textit{David} (Fig.~\ref{fig:capture})  scenes, we pass the laser through a diffuser; these scenes contain indirect diffuse reflections and the \textit{Kennedy} scene includes a mirror reflection.
Finally, the \textit{mirror} and \textit{diffraction} (see supplement) scenes are captured by illuminating a tank of water and milk with a collimated beam. 
In the \textit{water} scene the light passes through a mirror and is directed to a diffuse target.
The \textit{diffraction} scene captures a collimated beam passing through a diffraction grating. 

Overall, we observe similar trends for captured transient novel view synthesis as in simulated data.
However, because the captured data uses roughly 4$\times$ the number of histogram bins as simulated data (for finer the temporal resolution), we find that we need to increase the number of parameters in the K-planes model to reliably fit the data.
Due to the increased parameter count the performance of K-Planes improves upon the proposed method without modeling the propagation delay. 
However, this improvement comes at the cost of longer inference times, where K-Planes takes $80 \times$ longer to render a single transient video than the proposed method (see Table~\ref{tab:results}). 
Qualitatively, the proposed method recovers more plausible videos with fewer artifacts compared to the baselines (Fig.~\ref{fig:capture}). 
The approach also outperforms the baselines in terms of image quality and transient IoU (Table~\ref{tab:results}).
See the supplement for additional results and videos.

\subsection{Applications}
\label{sec:applications}

\begin{figure*}[t]
    \vspace{-1em}
    \centering
    \includegraphics[width=\textwidth]{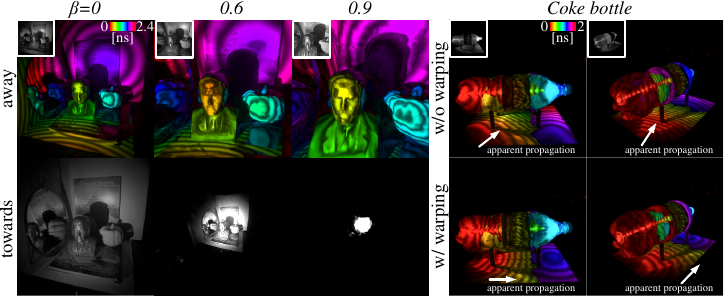}
    \captionof{figure}{Relativistic rendering and time warping. \textbf{Left:} We render relativistic effects due to motion towards and away from the scene at a fraction $\beta$ of the speed of light. Lorentz contraction~\cite{lorentz1892relative} causes objects to appear larger even though the camera moves away from them (top). Approaching the scene at a significant fraction of the speed of light results in increased brightness and shrinkage, resulting in the searchlight effect~\cite{jarabo2015relativistic} (bottom). \textbf{Right:} Peak-time visualizations for the \textit{Coke bottle} scene with and without depth-based warping. Transient appearance is consistent across viewpoints with depth-based warping.}
    \label{fig:combined}
    \vspace{-2em}
\end{figure*}

\paragraph{Time warping.} Our approach can be used to visualize light transport in different ways through time warping, which involves adding or removing time delays to the rendered transients. 
For example, following Velten et. al~\cite{velten2013femto} we can perform depth-based time warping, which removes the propagation delay from the scene point to the camera. 
In the visualization with depth-based warping, points along each camera ray ``light up'' as soon as they intersect a wavefront of light; in other words, each camera ray is rendered in a different spacetime coordinate frame. 

To create this visualization, we calculate the propagation delay to each scene point using the expected ray termination distance~\cite{deng2022depth}, and we shift the rendered transient by the corresponding speed-of-light time delay.
We compare transients rendered with and without depth-based warping in Fig.~\ref{fig:combined}. 
Here, the hue of the visualization indicates the time of peak intensity at each pixel (typically corresponding to the direct component of light).
The brightness is scaled by the maximum pixel intensity, and we modulate the brightness over the time dimension to add black bands corresponding to isochrones (wavefronts with equal path length). 
The appearance of the depth-warped transient is consistent across viewpoints, providing a more intuitive visualization.

Our representation enables more general time warping techniques, wherein we shift the transients based on the distance to arbitrary reference surfaces (e.g., defined by a sphere, cube, etc.). 
We explore extensions to time warping and associated novel visualization techniques in more detail in the supplement.

\paragraph{Relativistic rendering.} We consider rendering scene appearance from a camera moving at relativistic speeds. 
Rendering such effects has been explored in previous work~\cite{jarabo2015relativistic,weiskopf1999relativity} and we adapt these techniques into our transient rendering framework.
As an observer approaches the speed of light, different effects alter scene appearance as shown in Fig.~\ref{fig:combined}.
Specifically, we model (1) time dilation between the moving inertial frame of the camera and the static scene; (2) deformation of the camera focal length due to Lorentz contraction~\cite{lorentz1892relative,fitz1889ether}; (3) light aberration, which causes light rays to curve and compress towards the direction of motion; and (4) the searchlight effect, which describes the increase in photon flux for a camera traveling at relativistic speeds towards an illumination source. 
We provide implementation details and additional results in the supplement.

\begin{wrapfigure}[21]{r}{2.8in}
    \vspace{-5em}
    \includegraphics[height=2.9in,width=2.8in]{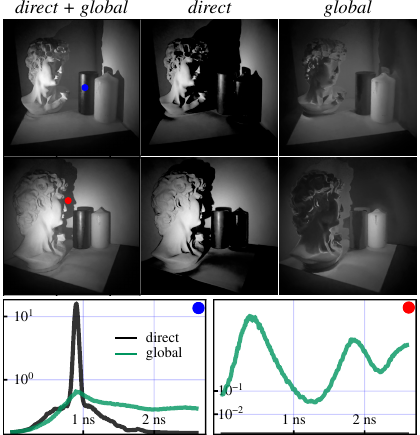}
    \caption{Visualization of the direct--global separation on the \textit{David} scene. The visualization of global illumination captures effects such as the interreflections on the wall or subsurface scattering in the candles.}
    \label{fig:separation}
\end{wrapfigure}

\paragraph{Direct--global separation.} We show how our method can be used for 3D visualization of direct and global components of light transport.
First, we pre-process the captured transient data to separate direct and global components by fitting a Gaussian mixture model to each transient.  
To identify the direct component, we check if the Gaussian closest to time zero matches the expected profile of the laser impulse used to illuminate the scene.
We model the global component using the remaining Gaussians (see the supplement for a detailed description of this procedure).
Finally, we train separate instances of our model on the direct and global components, allowing us to synthesize the corresponding direct and global transient videos (see Fig.~\ref{fig:separation}).

\section{Discussion}
Our work introduces a method for \textit{flying with photons}: rendering propagating light from novel, moving camera viewpoints.  
We envision many avenues for potential impact based on our technique, including in applications related to education, art, or for scientific observations of ultrafast phenomena that have so far been limited to capture from a single viewpoint. 
While the main focus of this work is on rendering and visualizing transient phenomena, multi-viewpoint transient videos are a rich source of scene information. 
We aim to develop new methods that use these data to infer scene geometry, reflectance, material properties, and more.
As such, we believe extensions of the proposed approach could have a broad array of applications in 3D computer vision, remote sensing, and biomedical imaging.
Due to the long acquisition times required for multi-view transient videography, our method is currently limited to static scenes. 
Acquisition times could potentially be reduced by simultaneous multi-view captures with emerging time stamping SPAD arrays, which may allow capturing transient effects within dynamic scenes.
We look forward to new advances based on multi-viewpoint transient videography.

\section*{Acknowledgments}
DBL and KNK acknowledge support of NSERC under the RGPIN, RTI, and Alliance programs. 
DBL also acknowledges support from the Canada Foundation for Innovation and the Ontario Research Fund.
RP acknowledges the support of the Stanford Graduate Fellowship program.

%
%
\bibliographystyle{splncs04}
\bibliography{refs}

\begin{thebibliography}{10}
\providecommand{\url}[1]{\texttt{#1}}
\providecommand{\urlprefix}{URL }
\providecommand{\doi}[1]{https://doi.org/#1}

\bibitem{abramson1978light}
Abramson, N.: Light-in-flight recording by holography. Opt. Lett.  \textbf{3}(4),  121--123 (1978)

\bibitem{ament2014radiative}
Ament, M., Bergmann, C., Weiskopf, D.: Refractive radiative transfer equation. ACM Trans. Graph.  \textbf{33}(2) (2014)

\bibitem{attal2021torf}
Attal, B., Laidlaw, E., Gokaslan, A., Kim, C., Richardt, C., Tompkin, J., O'Toole, M.: T{\"o}rf: Time-of-flight radiance fields for dynamic scene view synthesis. Proc. NeurIPS  \textbf{34} (2021)

\bibitem{bemana2022refractive}
Bemana, M., Myszkowski, K., Revall~Frisvad, J., Seidel, H.P., Ritschel, T.: Eikonal fields for refractive novel-view synthesis. In: Proc. ACM SIGGRAPH (2022)

\bibitem{bitterli16resources}
Bitterli, B.: Rendering resources (2016), https://benedikt-bitterli.me/resources/

\bibitem{blender}
{Blender Development Team}: Blender. \url{https://www.blender.org} (2023)

\bibitem{bronzi2015spad}
Bronzi, D., Villa, F., Tisa, S., Tosi, A., Zappa, F.: {SPAD} figures of merit for photon-counting, photon-timing, and imaging applications: a review. IEEE Sens. J.  \textbf{16}(1),  3--12 (2015)

\bibitem{cao2023hexplane}
Cao, A., Johnson, J.: Hexplane: A fast representation for dynamic scenes. In: Proc. CVPR (2023)

\bibitem{deng2022depth}
Deng, K., Liu, A., Zhu, J.Y., Ramanan, D.: Depth-supervised {NeRF}: Fewer views and faster training for free. In: Proc. CVPR (2022)

\bibitem{faccio2020non}
Faccio, D., Velten, A., Wetzstein, G.: Non-line-of-sight imaging. Nat. Rev. Phys.  \textbf{2}(6),  318--327 (2020)

\bibitem{fitz1889ether}
Fitz~Gerald, G.F.: The ether and the earth's atmosphere. Science (328),  390--390 (1889)

\bibitem{fridovich2023k}
Fridovich-Keil, S., Meanti, G., Warburg, F.R., Recht, B., Kanazawa, A.: {K-Planes}: Explicit radiance fields in space, time, and appearance. In: Proc. CVPR (2023)

\bibitem{gao2014single}
Gao, L., Liang, J., Li, C., Wang, L.V.: Single-shot compressed ultrafast photography at one hundred billion frames per second. Nature  \textbf{516}(7529),  74--77 (2014)

\bibitem{gariepy2015single}
Gariepy, G., Krstaji{\'c}, N., Henderson, R., Li, C., Thomson, R.R., Buller, G.S., Heshmat, B., Raskar, R., Leach, J., Faccio, D.: Single-photon sensitive light-in-fight imaging. Nat. Commun.  \textbf{6}(1), ~6021 (2015)

\bibitem{gkioulekas2015micron}
Gkioulekas, I., Levin, A., Durand, F., Zickler, T.: Micron-scale light transport decomposition using interferometry. ACM Trans. Graph.  \textbf{34}(4),  1--14 (2015)

\bibitem{gkioulekas2016evaluation}
Gkioulekas, I., Levin, A., Zickler, T.: An evaluation of computational imaging techniques for heterogeneous inverse scattering. In: Proc. ECCV (2016)

\bibitem{gupta2011stuctured}
Gupta, M., Agrawal, A., Veeraraghavan, A., Narasimhan, S.G.: Structured light {3D} scanning in the presence of global illumination. In: Proc. CVPR (2011)

\bibitem{heide2013low}
Heide, F., Hullin, M.B., Gregson, J., Heidrich, W.: Low-budget transient imaging using photonic mixer devices. ACM Trans. Graph.  \textbf{32}(4),  1--10 (2013)

\bibitem{hullin2008direct}
Hullin, M.B., Fuchs, M., Ajdin, B., Ihrke, I., Seidel, H.P., Lensch, H.P.: Direct visualization of real-world light transport. In: Proc. VMV (2008)

\bibitem{iseringhausen2020non}
Iseringhausen, J., Hullin, M.B.: Non-line-of-sight reconstruction using efficient transient rendering. ACM Trans. Graph.  \textbf{39}(1),  1--14 (2020)

\bibitem{jarabo2018bidirectional}
Jarabo, A., Arellano, V.: Bidirectional rendering of vector light transport. In: Computer Graphics Forum. vol.~37, pp. 96--105. Wiley Online Library (2018)

\bibitem{jarabo2014framework}
Jarabo, A., Marco, J., Munoz, A., Buisan, R., Jarosz, W., Gutierrez, D.: A framework for transient rendering. ACM Trans. Graph.  \textbf{33}(6),  1--10 (2014)

\bibitem{jarabo2017recent}
Jarabo, A., Masia, B., Marco, J., Gutierrez, D.: Recent advances in transient imaging: A computer graphics and vision perspective. Visual Informatics  \textbf{1}(1),  65--79 (2017)

\bibitem{jarabo2015relativistic}
Jarabo, A., Masia, B., Velten, A., Barsi, C., Raskar, R., Gutierrez, D.: Relativistic effects for time-resolved light transport. Computer Graphics Forum  \textbf{34},  1--12 (2015)

\bibitem{jin2023tenoir}
Jin, H., Liu, I., Xu, P., Zhang, X., Han, S., Bi, S., Zhou, X., Xu, Z., Su, H.: {TensoIR}: Tensorial inverse rendering. In: Proc., CVPR (2023)

\bibitem{kingma2015adam}
Kingma, D.P., Ba, J.: Adam: {A} method for stochastic optimization. In: Proc. ICLR (2015)

\bibitem{klinghoffer2023plato}
Klinghoffer, T., Xiang, X., Somasundaram, S., Fan, Y., Richardt, C., Raskar, R., Ranjan, R.: {PlatoNeRF}: {3D} reconstruction in {Plato's} cave via single-view two-bounce lidar. In: Proc. CVPR (2024), \url{https://platonerf.github.io}

\bibitem{kotwal2023passive}
Kotwal, A., Levin, A., Gkioulekas, I.: Passive micron-scale time-of-flight with sunlight interferometry. In: Proc. CVPR (2023)

\bibitem{li2023nerfacc}
Li, R., Gao, H., Tancik, M., Kanazawa, A.: {NerfAcc}: Efficient sampling accelerates {NeRFs}. arXiv preprint arXiv:2305.04966  (2023)

\bibitem{li2022nerfacc}
Li, R., Tancik, M., Kanazawa, A.: {NerfAcc}: A general {NeRF} acceleration toolbox. arXiv preprint arXiv:2210.04847  (2022)

\bibitem{li2022neural}
Li, T., Slavcheva, M., Zollhoefer, M., Green, S., Lassner, C., Kim, C., Schmidt, T., Lovegrove, S., Goesele, M., Newcombe, R., et~al.: Neural {3D} video synthesis from multi-view video. In: Proc. CVPR (2022)

\bibitem{liang2018single}
Liang, J., Wang, L.V.: Single-shot ultrafast optical imaging. Optica  \textbf{5}(9),  1113--1127 (2018)

\bibitem{lindell2018towards}
Lindell, D.B., O'Toole, M., Wetzstein, G.: Towards transient imaging at interactive rates with single-photon detectors. In: Proc. ICCP (2018)

\bibitem{liu2022temporally}
Liu, Y., Jiao, S., Jarosz, W.: Temporally sliced photon primitives for time-of-flight rendering. In: Computer Graphics Forum. vol.~41, pp. 29--40. Wiley Online Library (2022)

\bibitem{lorentz1892relative}
Lorentz, H.A.: The relative motion of the earth and the ether. Zittingsverlag Akad. V. Wet  \textbf{1},  74--79 (1892)

\bibitem{mai2023neural}
Mai, A., Verbin, D., Kuester, F., Fridovich-Keil, S.: Neural microfacet fields for inverse rendering (2023)

\bibitem{malik2023transient}
Malik, A., Mirdehghan, P., Nousias, S., Kutulakos, K.N., Lindell, D.B.: Transient neural radiance fields for lidar view synthesis and {3D} reconstruction. In: Proc. NeurIPS (2023)

\bibitem{matlabcamera}
Mathworks: Camera calibrator app. \url{https://www.mathworks.com/help/vision/ref/cameracalibrator-app.html} (2020)

\bibitem{max1995optical}
Max, N.: Optical models for direct volume rendering. IEEE Trans. Vis. Comput. Graph.  \textbf{1}(2),  99--108 (1995)

\bibitem{mildenhall2021nerf}
Mildenhall, B., Srinivasan, P.P., Tancik, M., Barron, J.T., Ramamoorthi, R., Ng, R.: {NeRF}: Representing scenes as neural radiance fields for view synthesis. Commun. ACM  \textbf{65}(1),  99--106 (2021)

\bibitem{mirdehghan2024turbosl}
Mirdehghan, P., Wu, M., Chen, W., Lindell, D.B., Kutulakos, K.N.: Turbosl: Dense accurate and fast {3D} by neural inverse structured light. In: Proc. CVPR (2024)

\bibitem{mu2024towards}
Mu, F., Sifferman, C., Jungerman, S., Li, Y., Han, M., Gleicher, M., Gupta, M., Li, Y.: Towards {3D} vision with low-cost single-photon cameras. In: Proc. CVPR (June 2024)

\bibitem{muller2022instant}
M{\"u}ller, T., Evans, A., Schied, C., Keller, A.: Instant neural graphics primitives with a multiresolution hash encoding. ACM Trans. Graph. (SIGGRAPH)  \textbf{41}(4),  1--15 (2022)

\bibitem{naik2011single}
Naik, N., Zhao, S., Velten, A., Raskar, R., Bala, K.: Single view reflectance capture using multiplexed scattering and time-of-flight imaging. ACM Trans. Graph. (SIGGRAPH Asia)  \textbf{30}(6),  1--10 (2011)

\bibitem{nayar2006fast}
Nayar, S., Krishnan, G., Grossberg, M., Raskar, R.: Fast separation of direct and global components of a scene using high frequency illumination. ACM Trans. Graph.  \textbf{25},  935--944 (07 2006)

\bibitem{oconnor1984time}
O'Connor, D.V., Phillips, D.: Time-correlated single photon counting. Academic Press (1984)

\bibitem{o2015homogeneous}
O'Toole, M., Achar, S., Narasimhan, S.G., Kutulakos, K.N.: Homogeneous codes for energy-efficient illumination and imaging. ACM Trans. Graph.  \textbf{34}(4),  1--13 (2015)

\bibitem{o2017reconstructing}
O'Toole, M., Heide, F., Lindell, D.B., Zang, K., Diamond, S., Wetzstein, G.: Reconstructing transient images from single-photon sensors. In: Proc. CVPR (2017)

\bibitem{o2014temporal}
O'Toole, M., Heide, F., Xiao, L., Hullin, M.B., Heidrich, W., Kutulakos, K.N.: Temporal frequency probing for {5D} transient analysis of global light transport. ACM Trans. Graph.  \textbf{33}(4),  1--11 (2014)

\bibitem{o20143d}
O'Toole, M., Mather, J., Kutulakos, K.N.: {3D} shape and indirect appearance by structured light transport. In: Proc. CVPR (2014)

\bibitem{o2012primal}
O'Toole, M., Raskar, R., Kutulakos, K.N.: Primal-dual coding to probe light transport. ACM Trans. Graph.  \textbf{31}(4),  39--1 (2012)

\bibitem{pan2022refractive}
Pan, J.I., Su, J.W., Hsiao, K.W., Yen, T.Y., Chu, H.K.: Sampling neural radiance fields for refractive objects. In: Proc. ACM SIGGRAPH Asia (2022)

\bibitem{pediredla2020path}
Pediredla, A., Chalmiani, Y.K., Scopelliti, M.G., Chamanzar, M., Narasimhan, S., Gkioulekas, I.: Path tracing estimators for refractive radiative transfer. ACM Trans. Graph.  \textbf{39}(6),  1--15 (2020)

\bibitem{pediredla2019ellipsoidal}
Pediredla, A., Veeraraghavan, A., Gkioulekas, I.: Ellipsoidal path connections for time-gated rendering. ACM Trans. Graph.  \textbf{38}(4),  1--12 (2019)

\bibitem{pifferi2016new}
Pifferi, A., Contini, D., Mora, A.D., Farina, A., Spinelli, L., Torricelli, A.: New frontiers in time-domain diffuse optics, a review. J. Biomed. Opt.  \textbf{21}(9),  091310--091310 (2016)

\bibitem{piron2020review}
Piron, F., Morrison, D., Yuce, M.R., Redout{\'e}, J.M.: A review of single-photon avalanche diode time-of-flight imaging sensor arrays. IEEE Sens. J.  \textbf{21}(11),  12654--12666 (2020)

\bibitem{rapp2017few}
Rapp, J., Goyal, V.K.: A few photons among many: Unmixing signal and noise for photon-efficient active imaging. IEEE Trans. Comput. Imaging  \textbf{3}(3),  445--459 (2017)

\bibitem{royo2022non}
Royo, D., García, J., Muñoz, A., Jarabo, A.: Non-line-of-sight transient rendering. Computers \& Graphics  \textbf{107},  84--92 (2022)

\bibitem{schonberger2016structure}
Schonberger, J.L., Frahm, J.M.: Structure-from-motion revisited. In: Proc. CVPR (2016)

\bibitem{shen2021non}
Shen, S., Wang, Z., Liu, P., Pan, Z., Li, R., Gao, T., Li, S., Yu, J.: Non-line-of-sight imaging via neural transient fields. IEEE Trans. Pattern Anal. Mach. Intell.  \textbf{43}(7),  2257--2268 (2021)

\bibitem{smith2008rendering}
Smith, A., Skorupski, J., Davis, J.: Transient rendering. Tech. Rep. UCSC-SOE-08-26, School of Engineering, University of California, Santa Cruz (February 2008)

\bibitem{tewari2022advances}
Tewari, A., Thies, J., Mildenhall, B., Srinivasan, P., Tretschk, E., Yifan, W., Lassner, C., Sitzmann, V., Martin-Brualla, R., Lombardi, S., et~al.: Advances in neural rendering. In: Computer Graphics Forum. vol.~41, pp. 703--735. Wiley Online Library (2022)

\bibitem{tsai2019beyond}
Tsai, C.Y., Sankaranarayanan, A.C., Gkioulekas, I.: Beyond volumetric albedo--a surface optimization framework for non-line-of-sight imaging. In: Proc. CVPR (2019)

\bibitem{velten2013femto}
Velten, A., Wu, D., Jarabo, A., Masia, B., Barsi, C., Joshi, C., Lawson, E., Bawendi, M., Gutierrez, D., Raskar, R.: Femto-photography: Capturing and visualizing the propagation of light. ACM Trans. Graph.  \textbf{32}(4), ~1--8 (2013)

\bibitem{verbin2022refnerf}
Verbin, D., Hedman, P., Mildenhall, B., Zickler, T., Barron, J.T., Srinivasan, P.P.: {Ref-NeRF}: Structured view-dependent appearance for neural radiance fields. In: Proc. CVPR (2022)

\bibitem{wang2023mixed}
Wang, F., Tan, S., Li, X., Tian, Z., Song, Y., Liu, H.: Mixed neural voxels for fast multi-view video synthesis. In: Proc. ICCV (2023)

\bibitem{wang2020single}
Wang, P., Liang, J., Wang, L.V.: Single-shot ultrafast imaging attaining 70 trillion frames per second. Nat. Commun.  \textbf{11}(1), ~2091 (2020)

\bibitem{wang2004image}
Wang, Z., Bovik, A.C., Sheikh, H.R., Simoncelli, E.P.: Image quality assessment: from error visibility to structural similarity. IEEE Trans. Image Process.  \textbf{13}(4),  600--612 (2004)

\bibitem{weiskopf2006explanatory}
Weiskopf, D., Borchers, M., Ertl, T., Falk, M., Fechtig, O., Frank, R., Grave, F., King, A., Kraus, U., Muller, T., et~al.: Explanatory and illustrative visualization of special and general relativity. IEEE Trans. Vis. Comput. Graph.  \textbf{12}(4),  522--534 (2006)

\bibitem{weiskopf1999relativity}
Weiskopf, D., Kraus, U., Ruder, H.: Searchlight and {Doppler} effects in the visualization of special relativity: A corrected derivation of the transformation of radiance. ACM Trans. Graph.  \textbf{18}(3),  278–292 (1999)

\bibitem{wu2021differentiable}
Wu, L., Cai, G., Ramamoorthi, R., Zhao, S.: Differentiable time-gated rendering. ACM Trans. Graph.  \textbf{40}(6),  1--16 (2021)

\bibitem{yi2021differentiable}
Yi, S., Kim, D., Choi, K., Jarabo, A., Gutierrez, D., Kim, M.H.: Differentiable transient rendering. ACM Trans. Graph.  \textbf{40}(6) (2021)

\bibitem{zhan2023nerfrac}
Zhan, Y., Nobuhara, S., Nishino, K., Zheng, Y.: {NeRFrac}: Neural radiance fields through refractive surface. In: Proc. ICCV (2023)

\bibitem{zhang2018unreasonable}
Zhang, R., Isola, P., Efros, A.A., Shechtman, E., Wang, O.: The unreasonable effectiveness of deep features as a perceptual metric. In: Proc. CVPR (2018)

\bibitem{zhang2021nerfactor}
Zhang, X., Srinivasan, P.P., Deng, B., Debevec, P., Freeman, W.T., Barron, J.T.: Nerfactor: Neural factorization of shape and reflectance under an unknown illumination. ACM Trans. Graph.  \textbf{40}(6),  1--18 (2021)

\bibitem{zhang2000flexible}
Zhang, Z.: A flexible new technique for camera calibration. IEEE Trans. Pattern Anal. Mach. Intell.  \textbf{22}(11),  1330--1334 (2000)

\end{thebibliography}

\includepdf[pages=-]{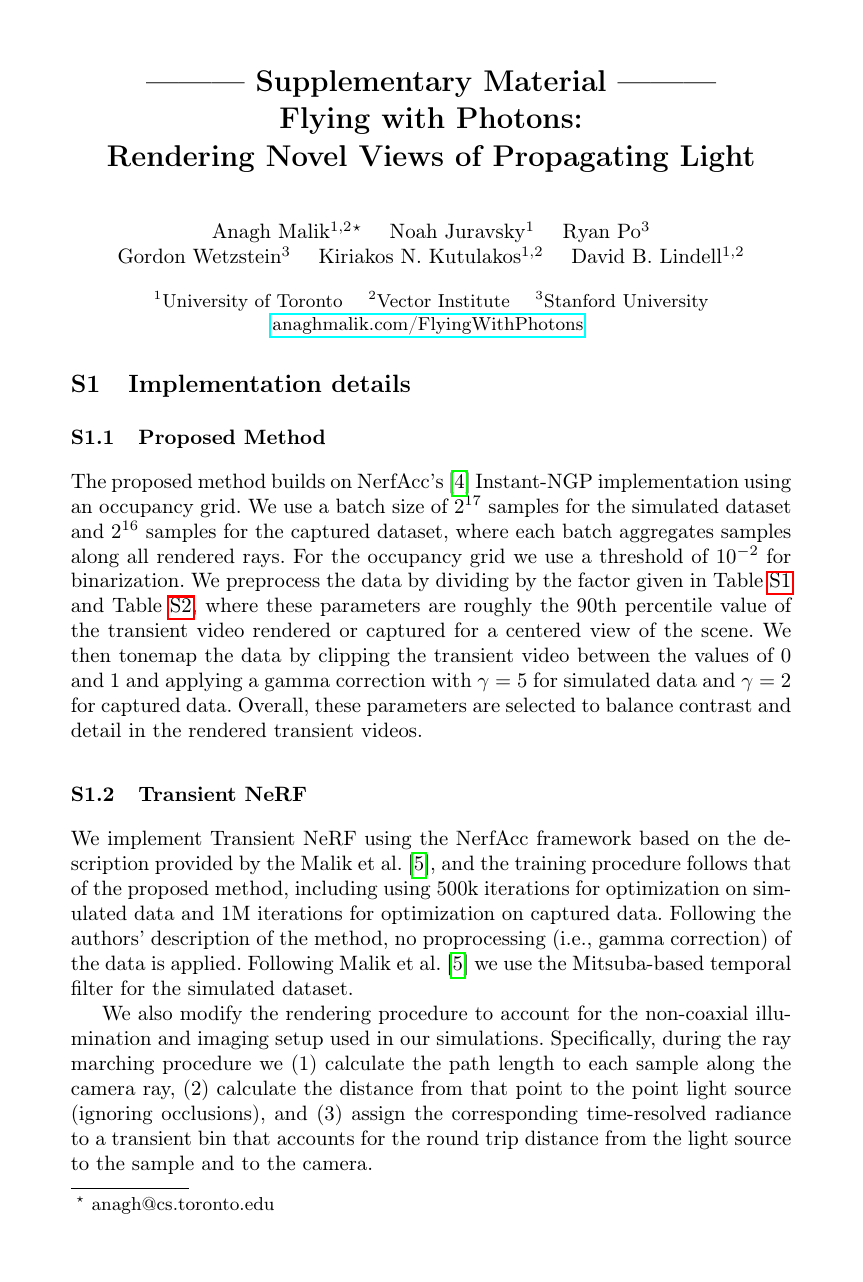}

\end{document}